\documentclass[fleqn,10pt,twocolumn,a4paper]{article}
\usepackage[a4paper,textwidth=180mm,textheight=242mm,columnsep=10mm,centering]{geometry}
\usepackage[T1]{fontenc}
\usepackage{times}
\usepackage{graphicx}
\usepackage{float}
\usepackage{array}
\usepackage{booktabs}
\usepackage{multirow}
\usepackage{makecell}
\usepackage{tikz}
\usetikzlibrary{bayesnet}
\usepackage{url}
\usepackage{changepage}
\usepackage{cite}
\usepackage[font=small,labelfont=bf,labelsep=period]{caption}
\usepackage[hidelinks]{hyperref}

\makeatletter
\renewcommand{\@seccntformat}[1]{\csname the#1\endcsname.\quad}
\renewcommand\section{\@startsection{section}{1}{\z@}{2.5ex plus .5ex minus .2ex}{1.5ex}{\normalfont\large\bfseries\raggedright\MakeUppercase}}
\renewcommand\subsection{\@startsection{subsection}{2}{\z@}{1.5ex}{.5ex}{\normalfont\normalsize\bfseries\raggedright}}
\renewcommand\subsubsection{\@startsection{subsubsection}{3}{\z@}{2ex plus .5ex minus .2ex}{.1ex}{\normalfont\normalsize\bfseries\raggedright}}
\makeatother
\newcommand{\argmax}{\mathop{\rm arg\,max}}
\newcommand{\argmin}{\mathop{\rm arg\,min}}
\hypersetup{pdftitle={Multimodal Person Concept Model for Person Identification under Varying Observation Conditions},pdfauthor={Mizuho Okitani, Yoshinobu Hagiwara}}
\begin{document}
\twocolumn[
\begin{center}
{\LARGE\bf Multimodal Person Concept Model for Person Identification\\[2pt]
under Varying Observation Conditions\par}
\vspace{1.0em}
{\large Mizuho Okitani and Yoshinobu Hagiwara\par}
\vspace{0.5em}
{\normalsize Faculty of Science and Engineering, Soka University, Tokyo, Japan\\
(E-mail: \href{mailto:e2358207@soka-u.jp}{e2358207@soka-u.jp}, \href{mailto:hagiwara@soka.ac.jp}{hagiwara@soka.ac.jp})\par}
\end{center}
\vspace{0.5em}
\begin{adjustwidth}{8mm}{8mm}
\small
\noindent\textbf{Abstract.}
Service robots in domestic and office environments must identify users to provide personalized services based on their preferences and previous interactions. This is challenging because only a few observations may be available for learning, while observation conditions change owing to appearance, environmental noise, and utterance content.
To address this challenge, this study proposes a multimodal person concept model that learns person concepts from a small number of observations, without person labels, and enables robust identification under changing conditions. The model employs a nonparametric Bayesian generative model based on the Chinese restaurant process and integrates visual, acoustic, and linguistic features extracted using Contrastive Language-Image Pre-training (CLIP) and Contrastive Language-Audio Pretraining (CLAP), followed by principal component analysis (PCA).
We evaluated six participants under eight conditions in which appearance changes, additive noise, and transcription errors were introduced individually and in combination. Averaged over the eight conditions, the proposed model achieved an identification accuracy of $0.89$ while inferring the number of person concepts from the observations, which is comparable to $k$-means clustering that was supplied with the true number of participants ($0.88$). We also examined PCA dimensionality, modality availability, and variation strength, and demonstrated the model on a Toyota Human Support Robot.
These results show that multimodal person concept learning supports robust person identification under changing conditions, and that identification can succeed even when the number of concepts is not correctly recovered.
The project page and demonstration video are
\href{https://okitamisan01.github.io/person_concept/}{available at https://okitamisan01.github.io/person\_concept/}.

\vspace{0.5em}
\end{adjustwidth}
\vspace{1.5em}
]

\section{Introduction}%
\label{sec:intro}%

For a domestic service robot to provide continuous and personalized services in homes, offices, or care facilities, it must identify users and respond to each of them individually.
For example, guiding a person or fetching an object based on a previous conversation is only possible when the robot can recognize who it is interacting with.
Therefore, person identification is an important capability for service robots that are expected to provide assistance based on users' preferences and previous interactions.

In real-world environments, however, person identification is challenging
because only a small number of observations may be available for learning.
At the same time, observation conditions can change substantially between
learning and identification.
For example, visual appearance may change because of clothing, posture,
or worn items; speech signals may be degraded by household and ambient
noise; and utterance content may vary between encounters.
As a result, a robot must identify users robustly from limited training
observations even when multiple modalities change simultaneously.

\begin{figure}[H]
\centering
\includegraphics[width=7.45cm]{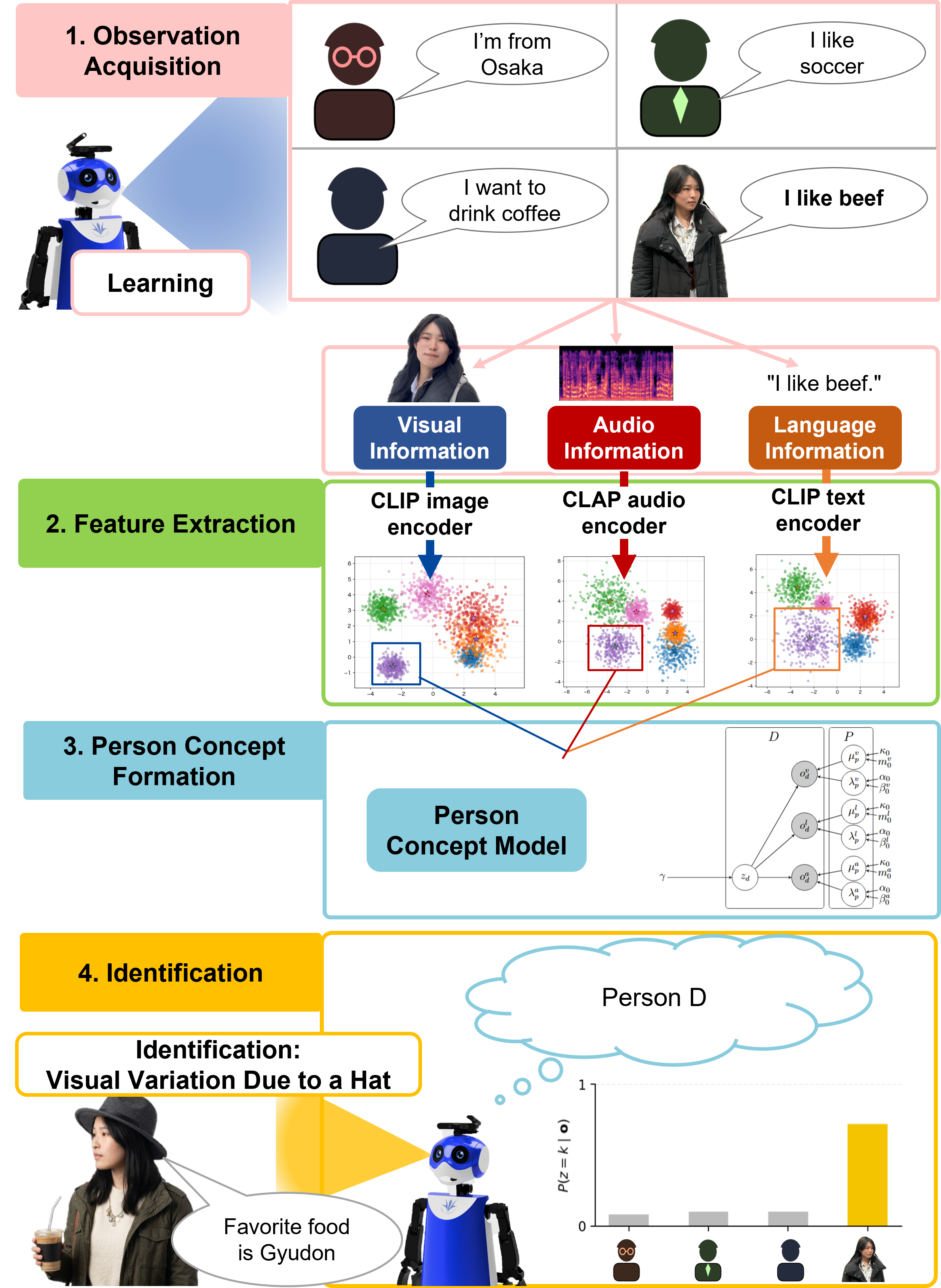}
\caption{\label{fig:intro} Overview of the proposed method.
Only the first author is shown.}
\end{figure}

To address this challenge, we propose a multimodal person concept model that learns person concepts from a small number of visual, acoustic, and linguistic observations.
An overview of this study is shown in Fig.~\ref{fig:intro}.
First, the robot collects visual, acoustic, and linguistic observations
of users in the environment. Next, visual and linguistic features are extracted using Contrastive
Language-Image Pre-training
\par\newpage
\newpage
\noindent (CLIP)~\cite{clip}, while acoustic features
are extracted using Contrastive Language-Audio Pretraining
(CLAP)~\cite{clap}, and these features are compressed using principal
component analysis (PCA).
The extracted multimodal features are then integrated by the
nonparametric Bayesian model based on the Chinese restaurant process
(CRP)~\cite{neal2000} to form person concepts.

Because the number of concepts is inferred from the observations rather
than specified in advance, the model can learn person concepts without
using person labels or prior knowledge of the number of users.
Finally, when a new observation is obtained, the model compares it with
the learned person concepts and identifies the person based on the
available modalities.
By integrating multiple modalities, the model can compensate for
degradation in one modality using information from the remaining
modalities, enabling person identification under changing observation
conditions.

The contributions of this study are as follows.

\vspace{0.25em}

\begin{adjustwidth}{1.25em}{0pt}

\begin{enumerate}
\item We develop a nonparametric multimodal person concept model that applies concept formation, previously used mainly for objects, motions, and places, to person identification, a setting that is distinct because observation conditions change dynamically between learning and identification. The model forms person concepts from only a small number of visual, acoustic, and linguistic observations, without person labels or prior knowledge of the number of users.

\item Through a systematic evaluation over all eight combinations of visual, acoustic, and linguistic variation, together with analyses of feature dimensionality, modality availability, and variation strength, we reveal that person concepts learned from limited observations remain useful for identification under changing conditions, and that accurate estimation of the number of concepts and accurate identification are separable: identification can succeed even when the number of concepts is not correctly recovered.

\item We demonstrate on a Toyota Human Support Robot that a learned person concept can be accessed from a single visual or acoustic observation and used to recall linguistic information from a previous interaction, illustrating how person concepts can support person-dependent service on a real robot.

\end{enumerate}

\end{adjustwidth}

\section{Related work}
\label{sec:related}

\subsection{Person identification from visual and acoustic cues}

Recent person-identification studies have widely used biometric
representations obtained with deep neural networks.
Liu et al.\ introduced iQIYI-VID, a large-scale multimodal dataset
containing substantial variations in pose, clothing, face quality, and
other observation conditions, and evaluated person identification using
face, body, and voice cues~\cite{iqiyi}.
Nagrani et al.\ learned cross-modal representations to match faces and
voices~\cite{nagrani2018}, while Brown et al.\ combined face, body, and
voice information for person clustering~\cite{brown2021}.
These studies demonstrate the effectiveness of multimodal biometric
information under diverse observation conditions, but generally rely on
large annotated datasets.

\par

To address limited enrollment data, Foggia et al.\ proposed a few-shot
speaker re-identification system for social robots using only three
reference voice samples per speaker~\cite{foggia2023}.
However, their method assumes speaker-specific enrollment, where
reference samples are associated with known speaker identities, and
focuses only on acoustic information.

\par

In contrast, our study aims to achieve person identification that is
robust to changes in observation conditions by learning from a small
number of multimodal observations acquired in the field, without using
person identity labels.

\subsection{Multimodal concept formation by robots}

In this context, probabilistic generative models offer a framework
for learning latent concepts from a small number of multimodal
observations without predefined class labels.
Nakamura et al.\ formed multimodal object concepts from visual, auditory,
and haptic observations~\cite{nakamura2009}, and later introduced a
hierarchical Dirichlet process to infer the number of categories from the
data~\cite{nakamura2011}.
Fadlil et al.\ further integrated object and motion concepts to acquire
action-level concepts without supervision~\cite{fadlil2013}.

\par

This framework has also been applied to spatial concepts.
Hagiwara et al.\ formed hierarchical spatial concepts using a nested
Chinese restaurant process~\cite{hagiwara2018}, while Taniguchi et al.
acquired spatial concepts and associated words during localisation and
mapping~\cite{spcoslam}.
These studies show that nonparametric Bayesian models can integrate
multimodal observations while inferring latent concepts and, in some
cases, the number of concepts from the observations themselves.

\par

However, previous multimodal concept-formation studies have mainly
focused on objects, motions, and places rather than people, whose
observations change between encounters.
A person concept formed from a few observations must therefore remain
accessible from observations that no longer resemble those it was formed
from.
Building on this line of work, we apply nonparametric multimodal concept
formation to person identification under this condition.

\section{Person concept model}
\label{sec:model}
As shown in Fig.~\ref{fig:intro}, the proposed model integrates features from the visual, acoustic and
linguistic modalities and aims to form one concept per person.  It consists of
four stages: observation acquisition, feature extraction, person concept
formation, and identification.
\subsection{Observation acquisition}
As shown in Fig.~\ref{fig:intro}(1), the robot stands about $1.0$~m in front of the participant and records a
video while the participant reads a self-introduction script. The video is divided into $10$~s clips,
indexed by $d=1,\ldots,D$, where $D$ denotes the total number of clips.  For each clip $d$, we take one representative frame as the visual
observation, the waveform of the whole clip as the acoustic observation,
and the transcription produced by Whisper as the linguistic observation.

\subsection{Feature extraction}
\label{sec:feature}
As shown in Fig.~\ref{fig:intro}(2), a $512$-dimensional visual feature $o^v_d$ is extracted from the frame with
CLIP (ViT-B/32)~\cite{clip}, a $512$-dimensional linguistic feature $o^l_d$ from the
transcription with the CLIP text encoder, and a $512$-dimensional
acoustic feature $o^a_d$ from the waveform with CLAP (htsat-fused)~\cite{clap}.
Principal component analysis (PCA) is then fitted on the training data
independently for each modality and reduces each $512$-dimensional
feature to four dimensions; the same projection is applied to the
identification data.  The compressed features are the observations of the
generative model.  

For the linguistic modality, the participants speak Japanese, so the
speech is transcribed by Whisper (large-v3) with the language set to
Japanese, and the resulting Japanese text is encoded using the CLIP text
encoder, whose training data consist primarily of English-language
image--text pairs.
The effect of this language mismatch on person concept formation is
examined in Section~\ref{sec:ablation}.

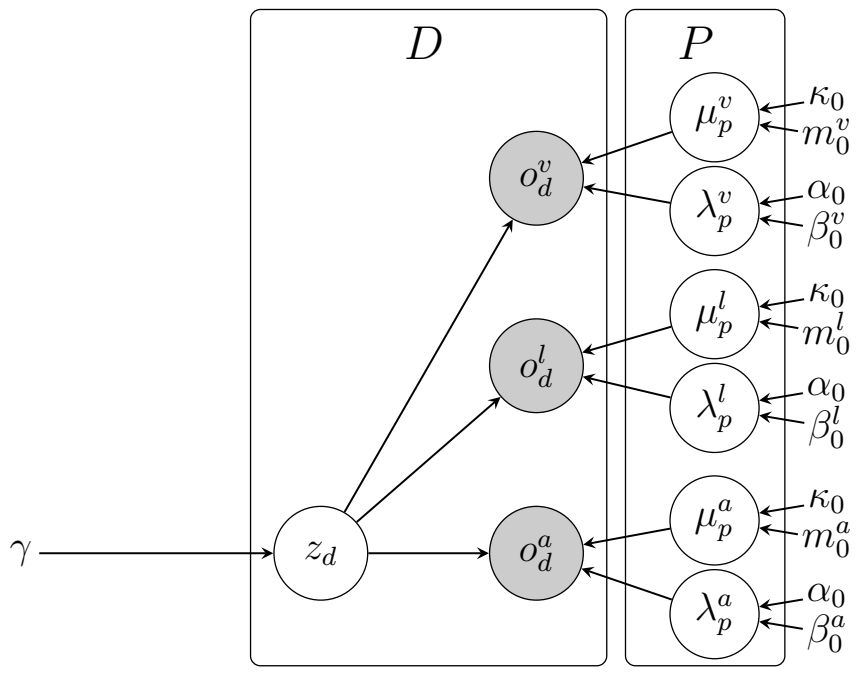
\begin{figure}[t]
\begin{center}
\scalebox{0.62}{%
\begin{tikzpicture}[
    every node/.style = {font=\large},
    hyper/.style  = {inner sep=2pt},
    obsW/.style   = {circle, draw, fill=white, minimum size=1.0cm, inner sep=0pt},
    obsG/.style   = {circle, draw, fill=gray!40, minimum size=1.0cm, inner sep=0pt},
    param/.style  = {circle, draw, fill=white, minimum size=0.95cm, inner sep=0pt},
    small/.style  = {inner sep=1pt, font=\large},
    factor/.style = {->, >=stealth, semithick},
  ]
  \node[hyper]  at (-0.7, -4.0) (gamma) {$\gamma$};
  \node[obsW]   at (2.5,  -4.0) (z)     {$z_d$};
  \node[obsG]   at (4.8,  0.0) (ov) {$o_d^v$};
  \node[obsG]   at (4.8, -2.0) (ol) {$o_d^l$};
  \node[obsG]   at (4.8, -4.0) (oa) {$o_d^a$};
  \node[param]  at (6.7,  0.65) (muV)  {$\mu_p^v$};
  \node[param]  at (6.7, -0.35) (lamV) {$\lambda_p^v$};
  \node[param]  at (6.7, -1.45) (muL)  {$\mu_p^l$};
  \node[param]  at (6.7, -2.45) (lamL) {$\lambda_p^l$};
  \node[param]  at (6.7, -3.65) (muA)  {$\mu_p^a$};
  \node[param]  at (6.7, -4.65) (lamA) {$\lambda_p^a$};
  \node[small] at (7.9,  0.85) (kapV) {$\kappa_0$};
  \node[small] at (7.9,  0.45) (m0V)  {$m^v_0$};
  \node[small] at (7.9, -0.15) (a0V)  {$\alpha_0$};
  \node[small] at (7.9, -0.55) (b0V)  {$\beta^v_0$};
  \node[small] at (7.9, -1.25) (kapL) {$\kappa_0$};
  \node[small] at (7.9, -1.65) (m0L)  {$m^l_0$};
  \node[small] at (7.9, -2.25) (a0L)  {$\alpha_0$};
  \node[small] at (7.9, -2.65) (b0L)  {$\beta^l_0$};
  \node[small] at (7.9, -3.45) (kapA) {$\kappa_0$};
  \node[small] at (7.9, -3.85) (m0A)  {$m^a_0$};
  \node[small] at (7.9, -4.45) (a0A)  {$\alpha_0$};
  \node[small] at (7.9, -4.85) (b0A)  {$\beta^a_0$};
  \draw[rounded corners=3pt] (1.75, 1.8) rectangle (5.55, -5.2);
  \node[font=\itshape\Large] at (3.6, 1.45) {$D$};
  \draw[rounded corners=3pt] (5.75, 1.8) rectangle (7.45, -5.2);
  \node[font=\itshape\Large] at (6.5, 1.45) {$P$};
  \draw[factor] (gamma) -- (z);
  \draw[factor] (z) -- (ov);
  \draw[factor] (z) -- (ol);
  \draw[factor] (z) -- (oa);
  \draw[factor] (muV)  -- (ov);  \draw[factor] (lamV) -- (ov);
  \draw[factor] (muL)  -- (ol);  \draw[factor] (lamL) -- (ol);
  \draw[factor] (muA)  -- (oa);  \draw[factor] (lamA) -- (oa);
  \draw[factor] (kapV) -- (muV); \draw[factor] (m0V) -- (muV);
  \draw[factor] (a0V)  -- (lamV);\draw[factor] (b0V) -- (lamV);
  \draw[factor] (kapL) -- (muL); \draw[factor] (m0L) -- (muL);
  \draw[factor] (a0L)  -- (lamL);\draw[factor] (b0L) -- (lamL);
  \draw[factor] (kapA) -- (muA); \draw[factor] (m0A) -- (muA);
  \draw[factor] (a0A)  -- (lamA);\draw[factor] (b0A) -- (lamA);
\end{tikzpicture}}
\caption{\label{fig:pgm} Graphical model of the proposed model.}
\end{center}
\end{figure}
\begin{table}[t]
\caption{\label{tab:variables} Definition of symbols used in the graphical model
($c \in \{v,l,a\}$ denotes the visual, linguistic, and acoustic modalities).}

\begin{center}
{
\renewcommand{\arraystretch}{1.25}
\begin{tabular}{@{}p{2.0cm}p{5.4cm}@{}}
\hline
Symbol & Definition \\
\hline
$d,\,D$ & clip index, number of clips \\
$p,\,P$ & concept index, number of concepts \\
$z$ & concept assigned to a clip (latent) \\
$\gamma$ & concentration parameter of the CRP \\
$o^c$ & observed feature of each modality \\
$\mu^c$ & mean of a concept in each modality \\
$\lambda^c$ & per-dimension precision of a concept \\
$\Sigma^c$ & diagonal covariance, $\Sigma^c={\rm diag}(1/\lambda^c)$ \\
\makecell[l]{$\kappa_0, m^c_0,$\\ $\alpha_0, \beta^c_0$} & Normal-Gamma hyperparameters \\

\hline
\end{tabular}
}
\end{center}
\end{table}

\subsection{Person concept formation}
\label{sec:formation}

As shown in Fig.~\ref{fig:intro}(3), person concepts are formed from the
multimodal observations extracted in the preceding steps.
The graphical model is shown in Fig.~\ref{fig:pgm}, and the symbols used
in the graphical model are defined in Table~\ref{tab:variables}.
The model assumes that $D$ multimodal observations
$\{(o^v_d, o^l_d, o^a_d)\}_{d=1}^{D}$ are generated from latent concept
assignments $z_d$ and per-concept distribution parameters.
Bayesian inference estimates the concept assignment of each clip and the
corresponding distribution parameters, thereby forming person concepts.
\par
This section describes the generative process, Bayesian inference,
initialisation procedure, and selection of the final sample.

\subsubsection{Generative process}
The generative process is defined as follows. Each observation is assumed to follow a Gaussian with diagonal
covariance, with a conjugate Normal-Gamma prior on the mean and the
precision of every dimension:
\begin{eqnarray}
  \{z_d\}_{d=1}^{D} &\sim& {\rm CRP}(\gamma), \label{eq:gen-z}\\
  \lambda^c_{p,j} &\sim& {\rm Gam}(\alpha_0, \beta^c_{0,j}), \label{eq:gen-lam}\\
  \mu^c_{p,j} &\sim& {\cal N}\!\left(m^c_{0,j}, (\kappa_0\lambda^c_{p,j})^{-1}\right), \label{eq:gen-mu}\\
  o^c_d &\sim& {\cal N}\!\left(\mu^c_{z_d}, \Sigma^c_{z_d}\right), \label{eq:gen-o}
\end{eqnarray}
where $c \in \{v,l,a\}$ and $j$ denotes the feature dimension after PCA
($j=1,\ldots,J$). ${\rm CRP}(\gamma)$ denotes the Chinese restaurant
process with concentration parameter $\gamma$, and
${\rm Gam}(\alpha,\beta)$ denotes the Gamma distribution parameterised
by shape $\alpha$ and rate $\beta$.
The covariance matrix is diagonal,
$\Sigma^c_p = {\rm diag}(1/\lambda^c_{p,1},\ldots,1/\lambda^c_{p,J})$.

\subsubsection{Bayesian inference}
Given the generative model above, the posterior distribution is
approximated by Gibbs sampling. The concept assignment
of clip $d$ is drawn from
\begin{eqnarray}
  \lefteqn{p(z_d{=}p \mid {\bf o}_d, {\bf z}_{-d}, \{\mu^c,\Sigma^c\})} \nonumber\\
  &\propto& p(z_d{=}p \mid {\bf z}_{-d})
            \cdot p({\bf o}_d \mid z_d{=}p, \mu^c_p, \Sigma^c_p),
  \label{eq:posterior-z}
\end{eqnarray}
where the first factor is the CRP assignment prior, which does not
require the number of concepts $P$ to be fixed in advance,
\begin{equation}
  p(z_d{=}p \mid {\bf z}_{-d}) \propto
  \left\{
  \begin{array}{ll}
    D_p^{\backslash d} & \mbox{(existing concept $p$)} \\
    \gamma             & \mbox{(new concept)} ,
  \end{array}
  \right.
\label{eq:crp}
\end{equation}
where $D_p^{\backslash d}$ denotes the number of clips currently assigned
to concept $p$, excluding clip $d$ itself.

The second factor is the observation likelihood, obtained under the
assumption of conditional independence between modalities:
\begin{eqnarray}
  \lefteqn{p({\bf o}_d \mid z_d{=}p, \mu^c_p, \Sigma^c_p)} \nonumber\\
  &=& {\cal N}(o^v_d; \mu^v_p, \Sigma^v_p)
      \cdot {\cal N}(o^l_d; \mu^l_p, \Sigma^l_p) \nonumber\\
  & & \cdot\, {\cal N}(o^a_d; \mu^a_p, \Sigma^a_p) .
  \label{eq:likelihood}
\end{eqnarray}
For a new concept, no sampled $\mu^c_p$ or $\Sigma^c_p$ exists.
Therefore, one set of parameters is drawn from the Normal-Gamma prior and
used to evaluate Eq.~(\ref{eq:likelihood}).
This corresponds to Algorithm~8, the auxiliary-parameter method, of
Neal~\cite{neal2000}.

\subsubsection{Initialisation}
To initialise the Gibbs sampler without specifying the number of people,
every clip is initially assigned to its own concept, so that the initial
number of concepts is $P=D$.
The number of participants is therefore not supplied as an initial value.  During sampling, clips are merged into
shared concepts and concepts that become empty are removed, so the number
of concepts decreases.  

The following two steps are repeated $T$ times.

\vspace{0.35em}

\begin{adjustwidth}{1.25em}{0pt}

\begin{enumerate}
\item \textbf{Sampling the concept assignment}

$z_d$ is drawn from the categorical distribution obtained by
normalising the product of Eq.~(\ref{eq:crp}) and
Eq.~(\ref{eq:likelihood}) over all concepts including a new one.
Empty concepts are removed and a concept is added whenever the new
concept is chosen, so $P$ changes over the iterations.
Only the mixing proportion $\theta$ is marginalised out and $z_d$ is
updated directly in the CRP form, that is, a collapsed Gibbs sampler;
$\mu^c_p$ and $\Sigma^c_p$ are \emph{not} marginalised out, and the
values sampled in Step~2 are used in Eq.~(\ref{eq:likelihood}).

\vspace{0.35em}
\item \textbf{Updating the distribution parameters}

$\mu^c_p$ and $\lambda^c_p$ are sampled per dimension from the
Normal-Gamma conjugate posterior given the features assigned to
concept $p$.
\end{enumerate}

\end{adjustwidth}

\subsubsection{Selection of the final sample}
The sampler returns a set of assignments rather than a single one, so a
representative sample has to be chosen. Because the aim of this study is
to form person concepts \emph{without} being given the number of people
or any person label, the selection criterion itself must not use the
ground truth. We therefore adopt the least-squares clustering criterion
of Dahl~\cite{dahl2006}.

Let ${z^{(t)}}$ denote the samples obtained after the burn-in period
$T_{\rm burn}$. For a pair of clips indexed by $d$ and $e$
($d,e=1,\ldots,D$), let $\delta^{(t)}
_{de}$ be $1$ if the two clips are
assigned to the same concept in sample $t$ and $0$ otherwise.
Let $\Pi_{de}$ denote the posterior probability that clips $d$ and $e$
are assigned to the same concept. It is estimated from the post-burn-in
samples as
\begin{equation}
\Pi_{de} =
\frac{1}{T-T_{\rm burn}}
\sum_{t=T_{\rm burn}+1}^{T} \delta^{(t)}_{de},
\label{eq:psm}
\end{equation}
and the sample adopted as the final model is the one whose co-clustering
pattern is closest to $\Pi$ in the least-squares sense:
\begin{equation}
t^{*} =
\argmin_t \sum_{d<e}
\left(\delta^{(t)}_{de}-\Pi_{de}\right)^2.
\label{eq:dahl}
\end{equation}
Here, $t^{*}$ denotes the iteration index of the sample selected as the
final model, and $z^{(t^{*})}$ is used as the final concept assignment.

This criterion depends only on the samples themselves, so the whole
procedure---initialisation, inference, and selection---uses no person
labels and no prior knowledge of the number of people.

\subsection{Identification}

As shown in Fig.~\ref{fig:intro}(4), the learned person concepts are used
to identify a newly observed person.
At identification time, the concept index is predicted from the learned
parameters $\{\mu^c_p, \Sigma^c_p\}$ using the maximum log-likelihood
criterion,
\begin{eqnarray}
\hat{z} &=& \argmax_p \Bigl[ \log{\cal N}(o^v; \mu^v_p, \Sigma^v_p) \nonumber\\ & & \qquad +\log{\cal N}(o^l; \mu^l_p, \Sigma^l_p) \nonumber\\ & & \qquad +\log{\cal N}(o^a; \mu^a_p, \Sigma^a_p) \Bigr] . \label{eq:identify} 
\end{eqnarray}
Because the training data are balanced across people, the concept prior
is treated as uniform.
For evaluation, the inferred concept indices are associated with the
ground-truth person labels by majority voting on the training data.

\subsubsection{Parameters used for identification}

The distribution parameters used in Eq.~(\ref{eq:identify}) are
determined from the final concept assignment $z^{(t^{*})}$ selected in
Section~\ref{sec:formation}.
Specifically, rather than using the sampled values of $\mu^c_p$ and
$\Sigma^c_p$ at iteration $t^{*}$, we recompute the corresponding
Normal-Gamma posterior from the observations assigned to each concept by
$z^{(t^{*})}$ and use its posterior mean.
Thus, the parameters used for identification are
\begin{equation}
\mu^c_{p} = m^c_{n,p}, \qquad
\Sigma^c_{p} = {\rm diag}\!\left(
\frac{\beta^c_{n,p,j}}{\alpha^c_{n,p}-1}
\right),
\label{eq:postmean}
\end{equation}
where $(\kappa^c_{n,p}, m^c_{n,p}, \alpha^c_{n,p},
\beta^c_{n,p})$ denote the parameters of the Normal-Gamma posterior for
concept $p$ computed under the selected assignment $z^{(t^{*})}$.

\section{Identification experiment}
\label{sec:exp}

\begin{table}[b]
\caption{\label{tab:exp} Experimental parameters.}
\begin{center}
\renewcommand{\arraystretch}{1.25}
\setlength{\tabcolsep}{4pt}

\begin{tabular}{@{}ll p{0.42\columnwidth}@{}}
\hline
Item & Value & Detail \\
\hline
Participants & 6 & four male, 
two female, \newline in their twenties \\
Recording conditions & 3 & reference (training),\newline similar, varied \\
Recording length & 180 sec & per participant and\newline per condition \\
Clips per recording & 18 & $10$~s each \\
Clips per condition & 108 & $6$ participants $\times$ $18$ clips \\
\hline
\end{tabular}
\end{center}
\end{table}

\begin{table*}[t]
\caption{\label{tab:dataset}
Multimodal information and the datasets used in the experiment.
The recordings are in Japanese; the linguistic row shows an English
translation of each transcript, and the features are computed from the
original Japanese text.}
\begin{center}
{
\renewcommand{\arraystretch}{1.2}
\begin{tabular}{@{}
    m{2.3cm}
    >{\raggedright\arraybackslash}m{4.3cm}
    >{\raggedright\arraybackslash}m{4.3cm}
    >{\raggedright\arraybackslash}m{4.3cm}
@{}}
\hline
\textbf{Modality}
  & \centering\arraybackslash
    \textbf{Reference set}\par
    (training)
  & \centering\arraybackslash
    \textbf{Similar set}\par
    (identification)
  & \centering\arraybackslash
    \textbf{Varied set}\par
    (identification) \\
\hline
\makecell[l]{Visual\\(author)}
  & \parbox[c][2.4cm][c]{\linewidth}{
      \centering
      \includegraphics[height=2.4cm,keepaspectratio]{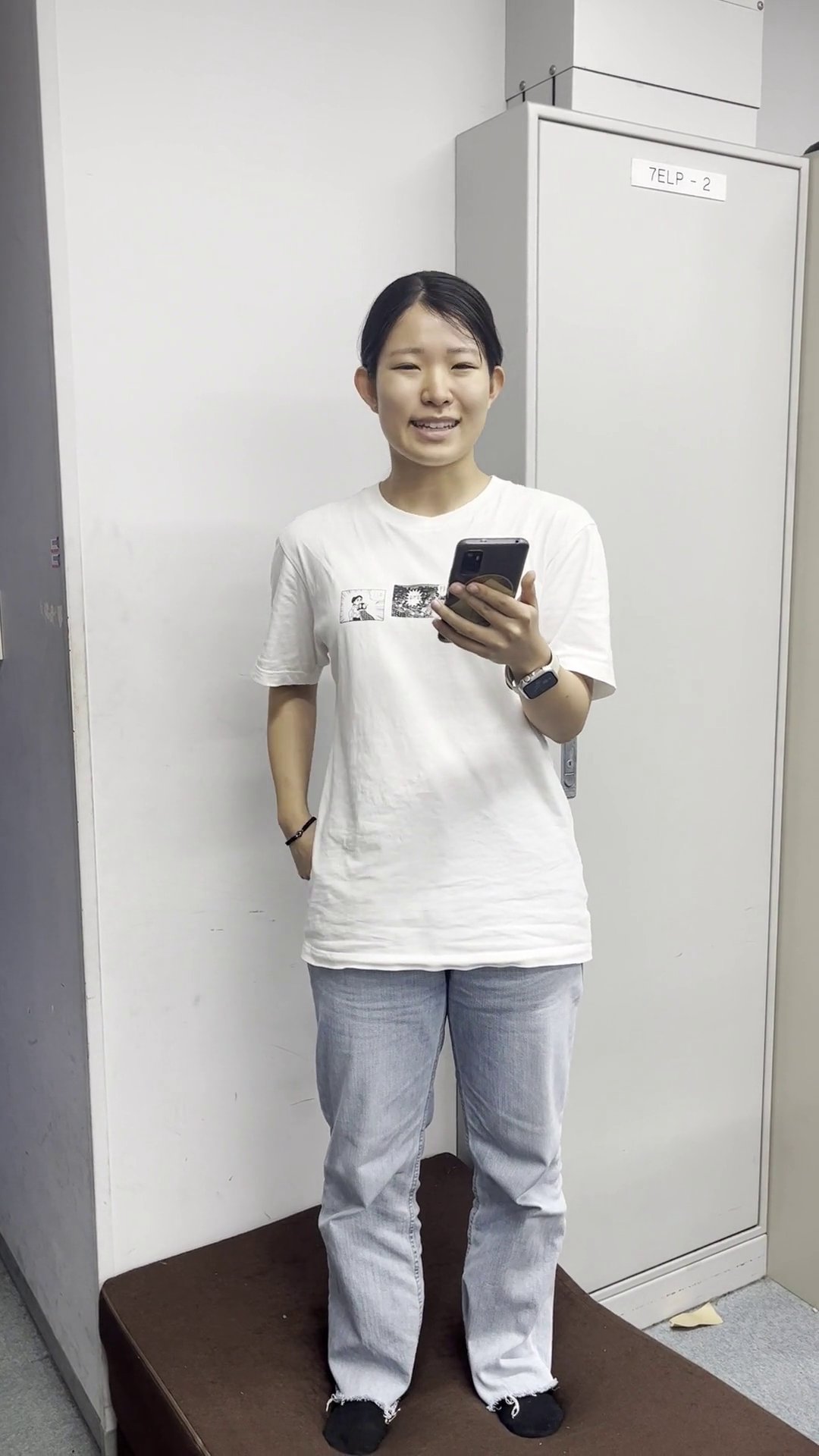}}
  & \parbox[c][2.4cm][c]{\linewidth}{
      \centering
      \includegraphics[height=2.4cm,keepaspectratio]{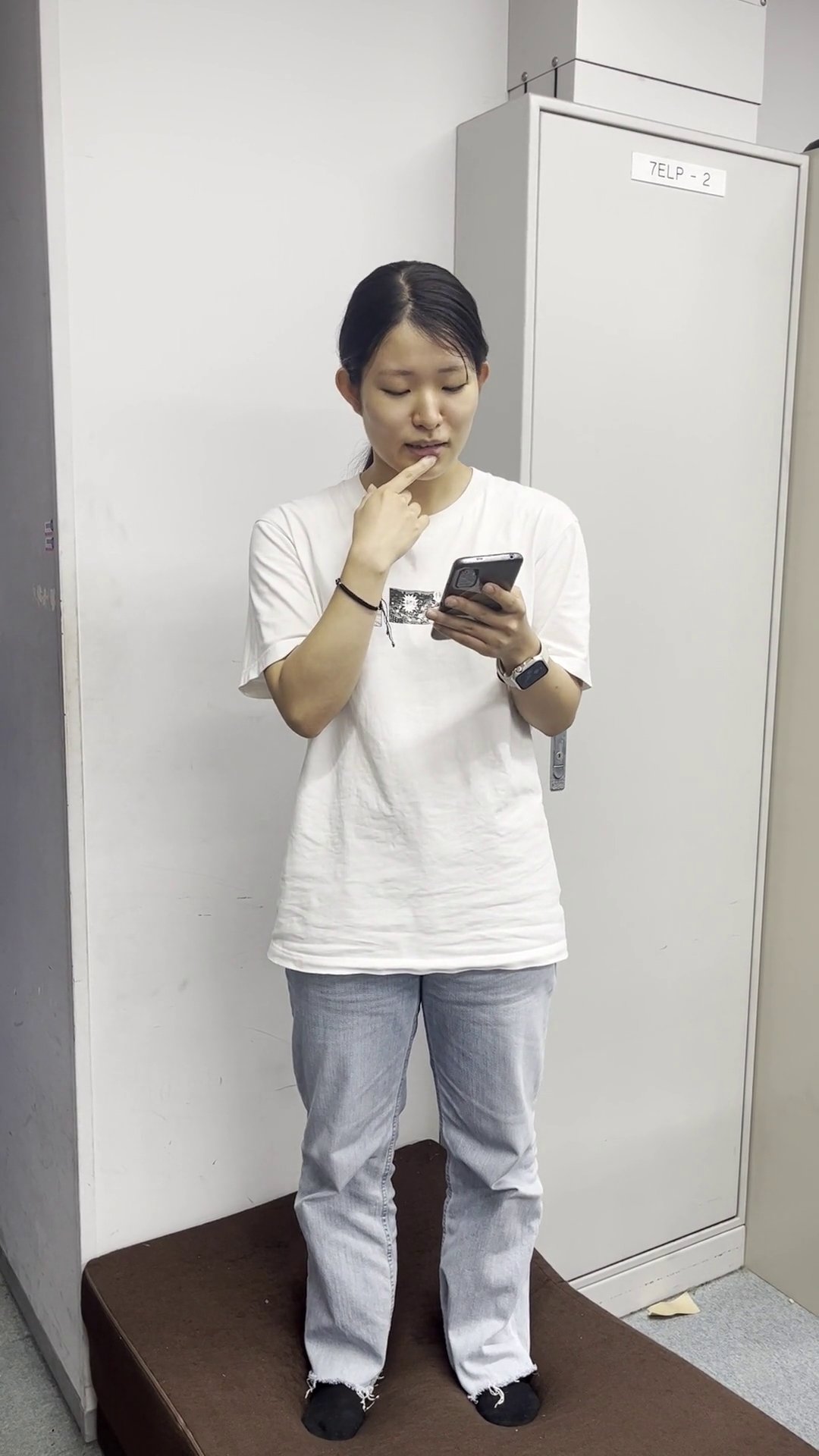}}
  & \parbox[c][2.4cm][c]{\linewidth}{
      \centering
      \includegraphics[height=2.4cm,keepaspectratio]{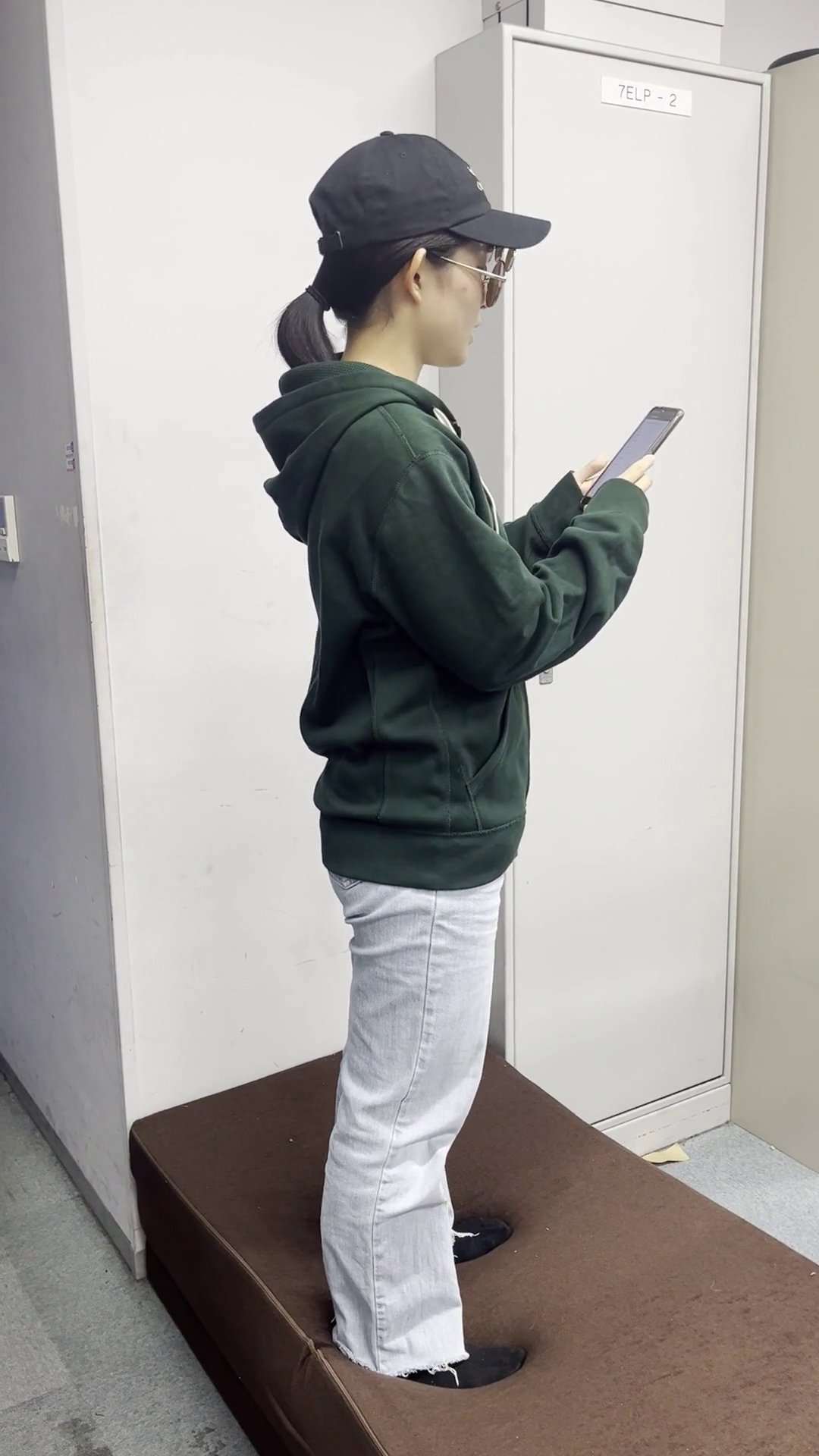}} \\
\hline
\makecell[l]{Acoustic\\(spectrogram)}
  & \parbox[c][1.5cm][c]{\linewidth}{
      \centering
      \includegraphics[width=2.3cm]{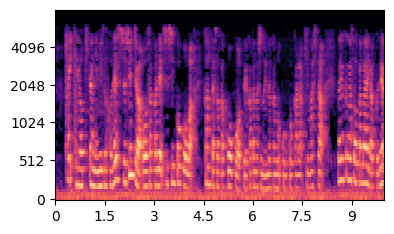}}
  & \parbox[c][1.5cm][c]{\linewidth}{
      \centering
      \includegraphics[width=2.3cm]{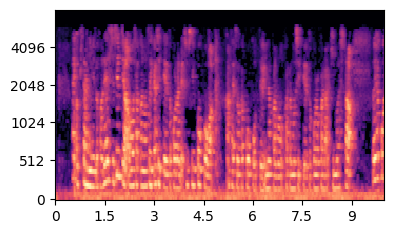}}
  & \parbox[c][1.5cm][c]{\linewidth}{
      \centering
      \includegraphics[width=2.3cm]{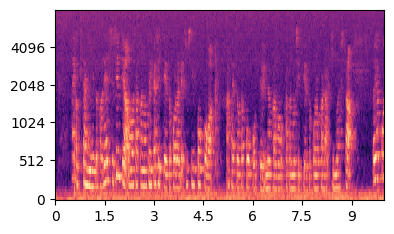}} \\
\hline
\makecell[l]{Linguistic\\(transcript,\\translated)}
  & I like beef. I basically like meat, and food that is Japanese-like.
  & As for favourite food, I like beef bowl. Japanese-like rice.
  & Recently I have been pushing Japanese-like eating. \\
\hline
\end{tabular}
}
\end{center}
\end{table*}

\subsection{Purpose}

We examine how variations in the observations of each modality affect
person-identification accuracy.
Eight conditions are defined by introducing visual, acoustic, and
linguistic variations individually and in combination.

\par

A further aim is to investigate whether person concepts can be formed
from a small number of observations without being given person identity
labels.
 As shown in Table~\ref{tab:exp}, the model is trained only on the reference condition, in which each
participant is represented by $18$ clips of $10$~s, and is then
evaluated on the similar and varied conditions without additional
training.
This setting reflects a situation in which a robot must acquire person
concepts from the limited interaction it can actually accumulate, rather
than from a large annotated corpus.

\par
In addition to the proposed nonparametric model, in which the number of
person concepts $P$ is inferred using the CRP, we evaluate the same
person-identification framework with $P$ fixed to $6$ and $7$.
These settings allow us to examine how fixing the number of concepts
affects identification performance compared with jointly inferring the
concept assignments and the number of concepts.

\par

To provide external comparisons, we also evaluate $k$-means clustering
with the true number of participants, $P=6$, and the pretrained
vision--language model Qwen2.5-VL.
The experiment therefore examines both robustness to changes in
observation conditions and the effect of inferring person concepts
without fixing their number in advance.

\subsection{Method}

\subsubsection{Data}

As shown in Table~\ref{tab:exp}, six participants (four male and two female, all in their twenties) were
recorded for approximately three minutes under each of three recording
conditions: reference, similar, and varied.

\par

As illustrated in the visual row of Table~\ref{tab:dataset}, the
reference condition consists of a self-introduction recorded in a stable
environment with the participant facing the camera.
The similar condition is a separate recording under the same stable
environment.
The varied condition includes appearance changes such as wearing
headphones, changing clothes, and turning the body.
Each video was divided into $10$~s clips.

\subsubsection{Conditions}

\begin{table}[t]
\caption{\label{tab:cond} Relation between the conditions and the datasets.}
\begin{center}
{
\renewcommand{\arraystretch}{1.25}
\begin{tabular}{@{}lccc@{}}
\hline
Condition & Visual & Acoustic & Linguistic \\
\hline
Reference (training) & ref. & ref. & ref. \\
\hline
Stable & sim. & sim. & sim. \\
Visual & var. & sim. & sim. \\
Acoustic & sim. & var. & sim. \\
Linguistic & sim. & sim. & var. \\
Visual + acoustic & var. & var. & sim. \\
Acoustic + linguistic & sim. & var. & var. \\
Linguistic + visual & var. & sim. & var. \\
All three & var. & var. & var. \\
\hline
\end{tabular}
}
\end{center}
\end{table}

The eight identification conditions shown in Table~\ref{tab:cond} were
evaluated.
In the varied acoustic condition, DKITCHEN noise from the DEMAND
dataset~\cite{demand} was superimposed at a signal-to-noise ratio (SNR)
of $10$~dB.
In the varied linguistic condition, each word in the transcription was
deleted or replaced with a corruption probability of $0.3$.

\par

The visual, acoustic, and linguistic variations were applied individually
and in combination to examine identification performance when one or
multiple modalities differ between learning and identification.

\begin{table*}[t]
\caption{\label{tab:result} Identification accuracy for each condition,
mean (standard deviation) over ten trials. The CRP model is not told how
many people there are: it inferred $P=6$ in six trials and $P=7$ in four
trials. All the other columns marked ``given'' are supplied with the
number of people; the $k$-means baseline is deterministic across the
trials. The three right-hand columns give the accuracy of Qwen2.5-VL from
the image only, the text only and both; it uses greedy decoding, so its
results do not vary over the trials.
Higher is better ($\uparrow$); in each row the best value is shown in
bold and underlined, and the second best in bold.}
\begin{center}
\renewcommand{\arraystretch}{1.25}
\setcellgapes{3pt}
\begin{tabular}{@{}p{4.2cm}ccccccc@{}}
\hline
\noalign{\vskip 2pt}
\multirow{2}{*}{Condition} & \multicolumn{3}{c}{Proposed method} & $k$-means & \multicolumn{3}{c}{Qwen2.5-VL (3B-Instruct)} \\[2pt]
\cline{2-4}\cline{6-8}
\noalign{\vskip 3pt}
 & \makecell{CRP\\[2pt]($P$ inferred) $\uparrow$}
 & \makecell{$P{=}7$\\[2pt]given $\uparrow$}
 & \makecell{$P{=}6$\\[2pt]given $\uparrow$}
 & \makecell{$P{=}6$\\[2pt]given $\uparrow$}
 & image $\uparrow$ & text $\uparrow$ & \makecell{image\\[2pt]+ text $\uparrow$} \\[3pt]
\hline
\noalign{\vskip 2pt}
Stable & ${\bf 0.98}\ (0.05)$ & $0.91\ (0.09)$ & $0.83\ (0.09)$ & $\underline{\bf 1.00}\ (0.00)$ & $0.36$ & $0.26$ & $0.27$ \\
Visual variation & ${\bf 0.85}\ (0.02)$ & $0.69\ (0.03)$ & $0.69\ (0.08)$ & $\underline{\bf 0.94}\ (0.00)$ & $0.22$ & $0.26$ & $0.28$ \\
Acoustic variation & $\underline{\bf 0.97}\ (0.06)$ & $0.70\ (0.07)$ & $0.69\ (0.16)$ & ${\bf 0.85}\ (0.00)$ & $0.36$ & $0.26$ & $0.27$ \\
Linguistic variation & ${\bf 0.98}\ (0.05)$ & $0.90\ (0.09)$ & $0.83\ (0.10)$ & $\underline{\bf 1.00}\ (0.00)$ & $0.36$ & $0.27$ & $0.28$ \\
Visual + acoustic variation & $\underline{\bf 0.75}\ (0.02)$ & $0.30\ (0.00)$ & $0.40\ (0.18)$ & $\underline{\bf 0.75}\ (0.00)$ & $0.22$ & $0.26$ & $0.28$ \\
Acoustic + linguistic variation & $\underline{\bf 0.97}\ (0.06)$ & $0.71\ (0.07)$ & $0.70\ (0.15)$ & ${\bf 0.84}\ (0.00)$ & $0.36$ & $0.21$ & $0.21$ \\
Linguistic + visual variation & ${\bf 0.85}\ (0.02)$ & $0.69\ (0.05)$ & $0.67\ (0.09)$ & $\underline{\bf 0.94}\ (0.00)$ & $0.22$ & $0.24$ & $0.25$ \\
Visual + acoustic + \newline linguistic variation & $\underline{\bf 0.76}\ (0.03)$ & $0.29\ (0.00)$ & $0.40\ (0.19)$ & ${\bf 0.72}\ (0.00)$ & $0.22$ & $0.25$ & $0.25$ \\
\noalign{\vskip 2pt}
\hline
\noalign{\vskip 2pt}
Average & $\underline{\bf 0.89}$ & $0.65$ & $0.65$ & ${\bf 0.88}$ & $0.29$ & $0.25$ & $0.26$ \\
\noalign{\vskip 2pt}
\hline
\end{tabular}
\end{center}
\end{table*}

\subsubsection{Metric}

Identification accuracy is defined as the proportion of correctly
identified clips among the $6 \times 18 = 108$ identification clips in
each condition.
For the proposed method, learning and inference are repeated ten times
on the same dataset, and the mean and standard deviation are reported.

\begin{figure}[t]
\begin{center}
\includegraphics[width=7cm]{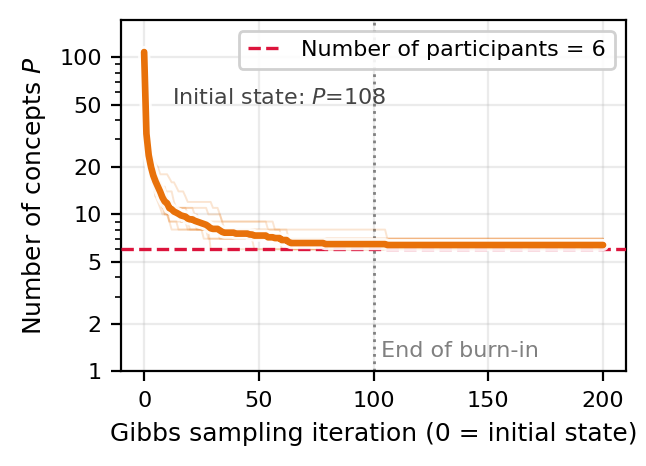}
\caption{\label{fig:convergence}
Number of concepts $P$ over the iterations in ten trials.
Iteration $0$ is the initial state in which every clip forms its own
concept ($P=D=108$).
The thick line represents the
mean over the ten trials.}
\end{center}
\end{figure}

\subsubsection{Comparison settings}

The proposed method uses the CRP described in
Section~\ref{sec:formation} to infer the number of person concepts from
the observations without being given the number of participants.

\par

To examine the effect of the number of concepts separately from the
nonparametric estimation of $P$, we additionally evaluate the proposed
person-concept model with the number of concepts fixed to $P=6$ and
$P=7$.
Except for fixing $P$, the same multimodal features and identification
procedure are used.

\par

To address whether the identification performance can be achieved by a
conventional clustering method when the correct number of people is
known, we compare the proposed method with $k$-means clustering with
$P=6$.
The same CLIP, CLAP, and PCA features and the same reference data are
used.
For $k$-means, each feature dimension is standardised using statistics
computed from the reference set because the modalities have different
scales.
The resulting clusters are associated with person labels by majority
voting on the training data.

\par

We also compare the proposed method with Qwen2.5-VL
(3B-Instruct)~\cite{qwen25vl}.
Qwen2.5-VL is evaluated using the image only, the
text only, and both the image and text.
Unlike the proposed method, Qwen2.5-VL does not learn person concepts
from the reference observations and does not use the acoustic modality.

\subsubsection{Model parameters}

For all modalities, $\kappa_0 = 0.01$ and $\alpha_0 = 2.0$.
The prior mean $m^c_0$ and rate parameter $\beta^c_0$ are determined
from the training features after PCA.
We set $\gamma = 0.75$, $T = 200$, and
$T_{\rm burn} = 100$.

\par

For the CRP setting, the singleton initialisation described in
Section~\ref{sec:formation} is used.
Because the training set contains $D=6\times18=108$ clips, the initial
number of concepts is $P=108$.
Thus, the number of participants is not supplied to the proposed
nonparametric model.

\begin{table*}[t]
\caption{\label{tab:dim}
Estimated number of concepts and identification performance for
different PCA dimensions (ten trials).
``Training ARI'' is the adjusted Rand index (ARI) between the inferred
partition and the participant labels on the training data.
``$P$ chg.'' is the number of the $199$ comparisons between consecutive
sweeps in which the number of concepts differed; the change from the
singleton initialisation to the first sweep is excluded, so a value of
$0.0$ means that the chain did not move after the first sweep.
It is a diagnostic of the sampler and has no preferred direction, whereas
$\uparrow$ marks the columns for which a larger value is better; the best
value in each such column is shown in bold and underlined.
The four-dimensional setting is the one used in Section~\ref{sec:exp}.}
\begin{center}
\renewcommand{\arraystretch}{1.35}
\begin{tabular}{@{}lcccc@{}}
\hline
PCA dimensions & Trials with $P{=}6$ $\uparrow$ & Training ARI $\uparrow$ &
$P$ chg.\ / $199$ & All var.\ $\uparrow$ \\
\hline
$3$ & $4/10$ & $0.963$ & $9.4$  & $0.75$ \\
$4$ & $\underline{\bf 6/10}$ & $0.961$ & $11.9$ & $\underline{\bf 0.76}$ \\
$5$ & $\underline{\bf 6/10}$ & $\underline{\bf 0.978}$ & $13.8$ & $0.68$ \\
$6$ & $4/10$ & $0.974$ & $14.3$ & $0.66$ \\
\hline
$512$ (no compression) & $0/10$ & $0.301$ & $0.0$ & $0.68$ \\
\hline
\end{tabular}
\end{center}
\end{table*}

\subsection{Results}

Table~\ref{tab:result} shows the identification accuracy under the eight
observation conditions.
The CRP model maintained high accuracy under single-modality variation,
achieving $0.97$ for acoustic variation and $0.98$ for linguistic
variation, compared with $0.98$ in the stable condition.
In contrast, visual variation reduced the accuracy to $0.85$.
A similar tendency was observed under multiple variations: the accuracy
was $0.97$ for acoustic + linguistic variation, but decreased to $0.75$
for visual + acoustic variation and $0.76$ when all three modalities
varied.

\par

To examine the effect of the concept number, we compared the CRP model
with the proposed model with $P$ fixed to $6$ and $7$.
The CRP model outperformed both fixed-$P$ settings in all eight
conditions, with an average accuracy of $0.89$ compared with $0.65$ for
both fixed-$P$ settings.
When all three modalities varied, the corresponding accuracies were
$0.76$, $0.40$, and $0.29$, respectively.

\par

To provide external comparisons, we further compared the CRP model with
$k$-means and Qwen2.5-VL.
The CRP model outperformed all Qwen2.5-VL settings in every condition.
Compared with $k$-means, which was given the true number of participants
($P=6$), the CRP model performed better under acoustic, acoustic +
linguistic, and all-three variation, while $k$-means performed better in
several other conditions.
Their average accuracies were $0.89$ and $0.88$, respectively, that is,
the CRP model matched the baseline on average even though the number of
participants was withheld from it and supplied to $k$-means.

\par

Figure~\ref{fig:convergence} shows the change in the number of concepts
during inference.
Starting from the singleton initialisation with $P=108$, the number of
concepts rapidly decreased and converged near the actual number of
participants by the end of the burn-in period.
The final number of concepts was $P=6$ in six trials and $P=7$ in four
trials.
Because the number of participants was not supplied to the CRP model,
these concept numbers were inferred from the multimodal observations.

\subsection{Discussion}

The results indicate that the proposed model is robust to changes in
observation conditions, although the effect differs across modalities.
Among the single-modality variations, visual variation caused a larger
decrease in accuracy than acoustic or linguistic variation.
Nevertheless, the CRP model achieved the highest accuracy among the
compared methods when all three modalities varied, indicating robustness
to combined observation changes.

\par

To address the role of concept-number inference, the CRP model was
compared with the same framework using fixed $P=6$ and $P=7$.
Although $P=6$ is the true number of participants, the CRP model
outperformed both fixed-$P$ settings in all conditions.
Together with the convergence from $P=108$ to approximately six
concepts, this suggests that jointly inferring the concept assignments
and the number of concepts is important for forming person concepts
useful for identification.
At the same time, the CRP model inferred $P=7$ in four of the ten trials
yet still reached an average accuracy of $0.89$, which indicates that
recovering the exact number of concepts and identifying people correctly
are not the same thing.
Section~\ref{sec:sensitivity} examines this separation directly.

\par

The baseline comparisons further support this interpretation.
The CRP model substantially outperformed Qwen2.5-VL, which does not
learn person-specific concepts from the reference observations.
Although $k$-means performed better in several conditions, it was given
the true number of participants in advance.
In contrast, the CRP model achieved comparable overall accuracy while
simultaneously estimating the number of person concepts from the
observations.

\par

Overall, these results support the feasibility of forming person concepts
from a small number of multimodal observations without person identity
labels or prior knowledge of the number of people.
However, the dataset contains only six participants, and the acoustic
and linguistic variations were synthetically introduced.
Evaluation with more participants and naturally occurring observation
changes is therefore required in future work.

\begin{table*}[t]
\caption{
Identification accuracy when the available modalities are restricted
(mean over ten trials).
Five representative conditions are shown.
}
\label{tab:ablation}
\begin{center}
{
\renewcommand{\arraystretch}{1.20}
\setlength{\extrarowheight}{2pt}
\begin{tabular}{@{}lccccccc@{}}
\hline
\multirow{2}{*}[-0.7ex]{Condition}
& \multicolumn{7}{c}{Modalities used for identification $\uparrow$} \\[3pt]
\cline{2-8}
\rule{0pt}{3.4ex}%
& Visual
& Acoustic
& Linguistic
& \makecell{Visual +\\[3pt]Acoustic}
& \makecell{Visual +\\[3pt]Linguistic}
& \makecell{Acoustic +\\[3pt]Linguistic}
& \makecell{Visual + Acoustic\\[3pt]+ Linguistic} \\[3pt]
\hline
\rule{0pt}{3ex}%
Stable
& $0.98$
& $0.89$
& $0.18$
& $0.98$
& $0.98$
& $0.89$
& $0.98$ \\
Visual var.
& $0.79$
& $0.89$
& $0.18$
& $0.86$
& $0.79$
& $0.89$
& $0.85$ \\
Acoustic var.
& $0.98$
& $0.53$
& $0.18$
& $0.97$
& $0.98$
& $0.59$
& $0.97$ \\
Linguistic var.
& $0.98$
& $0.89$
& $0.17$
& $0.98$
& $0.98$
& $0.89$
& $0.98$ \\
All var.
& $0.77$
& $0.53$
& $0.18$
& $0.73$
& $0.77$
& $0.58$
& $0.76$ \\
\hline
\end{tabular}
}
\end{center}
\end{table*}

\begin{table}[t]
\caption{
Effect of the modalities used for concept formation.
ARI is computed on the training data.
The $P=6$ column reports the number of trials in which the inferred
number of concepts matched the six participants.
}
\label{tab:formation}
\begin{center}
{
\renewcommand{\arraystretch}{1.25}
\setlength{\tabcolsep}{1.5pt}
\begin{tabular}{@{}lcc@{}}
\hline
Modalities & $P{=}6$ trials $\uparrow$ & Training ARI $\uparrow$ \\
\hline
Visual
& $3/10$
& $0.87$ \\
Acoustic
& $7/10$
& $0.89$ \\
Linguistic
& $5/10$
& $-0.01$ \\
Visual + Acoustic
& $10/10$
& $1.00$ \\
Visual + Linguistic
& $4/10$
& $0.80$ \\
Acoustic + Linguistic
& $0/10$
& $0.23$ \\
Visual + Acoustic + Linguistic
& $6/10$
& $0.96$ \\
\hline
\end{tabular}
}
\end{center}
\end{table}

\section{Additional experiment 1: Dimensionality}
\label{sec:sensitivity}
\subsection{Purpose}
We examine how feature dimensionality affects the estimation of person
concepts and identification performance.
We also examine whether correctly recovering the number of participants
is necessary for identification.
\par
We therefore compare the inferred number of concepts, training ARI, and
identification accuracy across different feature dimensions.

\subsection{Method}
Using the same dataset and settings as in Section~\ref{sec:exp}, we
varied the feature dimensionality and repeated learning ten times for
each setting.
We report the number of trials with $P=6$, training ARI, and
identification accuracy when all three modalities vary.
We also report the number of iterations in which $P$ changes
(``$P$ chg.'') to examine the behaviour of the sampler.

\subsection{Results}
Table~\ref{tab:dim} shows that dimensionality strongly affected concept
formation.
At four dimensions, $P=6$ was recovered in $6/10$ trials and the
training ARI was $0.961$.
The all-variation identification accuracy was $0.76$, the highest among
the evaluated settings.
\par
To address whether correct concept-number estimation is necessary for
identification, we also evaluated the original $512$-dimensional
features.
Although $P=6$ was recovered in $0/10$ trials and the training ARI
decreased to $0.301$, identification accuracy remained $0.68$.

\subsection{Discussion}
These results indicate that dimensionality reduction is important for
estimating the number of person concepts.
At high dimensions, large likelihood differences make concept
reassignment difficult, preventing concepts from being merged properly.
\par
However, accurate estimation of the number of concepts is not
necessarily required for identification.
Even at $512$ dimensions, where the model formed more concepts than the
number of participants, identification remained possible.
This is also consistent with the HSR demonstration, in which a person
could be identified and associated linguistic information could be
retrieved despite overestimating the number of concepts.
Thus, when the primary objective is person identification and
person-specific service, the exact number of learned concepts may be
less important.
\par
We therefore use four PCA dimensions in Section~\ref{sec:exp}, which
achieved both a high $P=6$ recovery rate and the highest identification
accuracy.

\section{Additional experiment 2: Modality ablation}
\label{sec:ablation}

\subsection{Purpose}

Section~\ref{sec:exp} showed that the proposed method maintained high
identification accuracy even when one or more modalities varied.
However, those results do not directly show whether this robustness
arises from complementary information across modalities.
\par
To address this gap, we conduct two ablation analyses.
First, we restrict the modalities available at identification time to
examine how different observations access an already formed person
concept.
Second, we retrain the model with different modality subsets to examine
how each modality contributes to concept formation.

\subsection{Method}

\subsubsection{Identification-time ablation}

The ten models learned in Section~\ref{sec:exp} are reused without
retraining.
Identification is repeated using each of the seven non-empty subsets of
the visual, acoustic, and linguistic modalities.
For each subset, only the corresponding log-likelihood terms in
Eq.~(\ref{eq:identify}) are used.
The seven subsets are evaluated under the eight conditions in
Table~\ref{tab:cond}, and the results are averaged over ten trials.
Thus, differences in accuracy reflect only the information available at
identification time.
\par

\subsubsection{Formation-time ablation}

The identification-time analysis does not show how each modality
contributes to forming the person concepts because all models were
originally learned from three modalities.
To examine this separately, the model is retrained from scratch using
each of the seven non-empty modality subsets.
The same hyperparameters, singleton initialisation, and label-free
selection rule in Eq.~(\ref{eq:dahl}) are used.
Concept formation is evaluated using the inferred number of concepts
and the ARI computed on the training data.

\subsection{Results}

Table~\ref{tab:ablation} shows the identification-time ablation.
When the varied modality was used alone, identification accuracy
decreased, whereas using all three modalities generally maintained
higher performance.
For example, under acoustic variation, accuracy increased from $0.53$
with Acoustic alone to $0.97$ with all three modalities.
Even when all three modalities varied, the three-modality configuration
maintained an accuracy of $0.76$.
\par
Table~\ref{tab:formation} shows the formation-time ablation.
Visual + Acoustic provided the most stable concept formation, recovering
$P=6$ in all ten trials with an ARI of $1.00$.
In contrast, adding Linguistic did not improve the ARI, and Linguistic
alone produced an ARI close to zero despite recovering $P=6$ in half of
the trials.

\subsection{Discussion}

The identification-time results indicate that the modalities provide
complementary information.
When one modality becomes unreliable, another modality can still support
access to the same person concept.
This explains why multimodal identification generally remained more
robust than identification based only on the varied modality.
\par
However, using all available modalities was not always optimal.
Under acoustic variation, excluding the degraded acoustic observation
slightly improved identification.
Because Eq.~(\ref{eq:identify}) weights the modality likelihoods equally,
an unreliable modality can reduce the fused likelihood.
Reliability-weighted multimodal fusion therefore remains an important
direction for future work.
\par
The formation-time results show that Visual and Acoustic provide the
main information required to form person concepts in this dataset.
By contrast, the present Linguistic representation contributes little
to separating persons.
The discrepancy between the inferred number of concepts and the ARI
also shows that obtaining the correct value of $P$ does not necessarily
mean that the observations are grouped by the correct persons.
\par
This does not imply that linguistic information is generally ineffective
for person modelling.
Rather, the current CLIP text representation of Japanese transcripts
appears to provide limited discriminative information in this dataset.
Two factors may contribute to this result.
First, as described in Section~\ref{sec:feature}, Japanese transcripts
are encoded using the CLIP text encoder, although CLIP was trained
primarily on English-language image--text pairs.
This language mismatch may therefore limit the informativeness of the
resulting linguistic features.
Second, the CLIP text representation is designed to capture semantic
content that can be aligned with images rather than information about
the identity of the speaker.
Because the participants answered similar questions, their utterances
may also have similar semantic content, making them difficult to
distinguish using these features alone.
\par
These results suggest a division of roles for the linguistic modality.
Rather than letting the linguistic feature participate in the clustering,
person concepts may be formed from the visual and acoustic modalities
alone, with the raw transcript of each clip simply associated with the
concept its observation was assigned to.
The linguistic information is then used for memory recall instead of for
separating persons, which is how the HSR demonstration in
Section~\ref{sec:demo} stores a representative utterance with each
concept.
Evaluating this configuration quantitatively, together with a linguistic
representation that could contribute to concept formation as well---for
example a multilingual or speaker-discriminative encoder---remains future
work.

\section{Additional experiment 3: Variation strength}
\label{sec:strong}

\subsection{Purpose}

In Section~\ref{sec:exp}, identification accuracy remained high even
under acoustic and linguistic variation.
However, this may be because the applied variation was too weak, rather
than because the remaining modalities compensated for the degraded
observation.
\par
To address this gap, we increase the strength of the acoustic and
linguistic variations and compare identification using the varied
modality alone with identification using all three modalities.
We further evaluate the linguistic variation end to end by
re-transcribing noisy speech.

\subsection{Method}

For acoustic variation, DKITCHEN noise was added at SNRs of $5$, $0$,
and $-5$~dB, in addition to the $10$~dB condition used in
Section~\ref{sec:exp}.
For linguistic variation, the synthetic word corruption probability was increased
from $0.3$ to $0.5$ and $0.8$.
The ten learned models from Section~\ref{sec:exp} were reused without
retraining, and only identification was repeated.
As in Section~\ref{sec:ablation}, we compare the varied modality alone
with identification using all three modalities.
\par
To evaluate a more realistic degradation of the linguistic observation,
the noisy waveforms were also transcribed using Whisper large-v3.
Because the transcripts are Japanese, degradation is measured by the
character error rate (CER) relative to the transcript obtained from the
same clean waveform.

\begin{table*}[t]
\caption{\label{tab:strong}
Identification accuracy under different strengths of acoustic and
linguistic variation, shown as mean (standard deviation) over ten
trials.
The strength is given as the SNR for the acoustic variation and as the
synthetic word corruption probability, that is, the probability with
which each word of the transcript is deleted or replaced, for the
linguistic variation.
``Varied modality alone'' uses only the modality that is degraded, that
is, the acoustic modality in the upper block and the linguistic modality
in the lower block.
The chance level for six-way classification is $0.17$.
The best value in each row is shown in bold and underlined.}
\begin{center}
{
\renewcommand{\arraystretch}{1.25}
\begin{tabular}{@{}llcc@{}}
\hline
Variation & Strength
& Varied modality alone $\uparrow$
& All three modalities $\uparrow$ \\
\hline
\multirow{4}{*}{\makecell[l]{Acoustic variation\\(SNR)}}
& $10$~dB & $0.53\ (0.06)$ & $\underline{\mathbf{0.97}}\ (0.06)$ \\
& $5$~dB  & $0.43\ (0.06)$ & $\underline{\mathbf{0.98}}\ (0.05)$ \\
& $0$~dB  & $0.39\ (0.05)$ & $\underline{\mathbf{0.98}}\ (0.05)$ \\
& $-5$~dB & $0.34\ (0.00)$ & $\underline{\mathbf{0.98}}\ (0.05)$ \\
\hline
\multirow{3}{*}{\makecell[l]{Linguistic variation\\(word corruption)}}
& $0.3$ & $0.17\ (0.02)$ & $\underline{\mathbf{0.98}}\ (0.05)$ \\
& $0.5$ & $0.19\ (0.01)$ & $\underline{\mathbf{0.98}}\ (0.05)$ \\
& $0.8$ & $0.16\ (0.02)$ & $\underline{\mathbf{0.98}}\ (0.05)$ \\
\hline
\end{tabular}
}
\end{center}
\end{table*}

\begin{table}[t]
\caption{\label{tab:e2e}
End-to-end degradation of the linguistic observation.
The noisy waveforms are transcribed again using Whisper large-v3, and the
resulting text is used as the linguistic observation; the visual and
acoustic observations are those of the corresponding acoustic condition.
Degradation is measured by the relative character error rate (CER)
against the transcript that the same procedure produces from the clean
waveform, so the clean condition is $0.00$ by definition; the values are
means over the $108$ clips.
Identification accuracy is computed using all three modalities and is
shown as mean (standard deviation) over the ten models learned in
Section~\ref{sec:exp}.}
\begin{center}
{
\renewcommand{\arraystretch}{1.25}
\begin{tabular}{@{}lcc@{}}
\hline
Acoustic condition & Relative CER & All three modalities $\uparrow$ \\
\hline
Clean
& $0.00$
& $0.98\ (0.05)$ \\
SNR $=10$~dB
& $0.15$
& $0.98\ (0.05)$ \\
SNR $=5$~dB
& $0.20$
& $0.98\ (0.05)$ \\
SNR $=0$~dB
& $0.30$
& $0.98\ (0.05)$ \\
SNR $=-5$~dB
& $0.53$
& $0.98\ (0.05)$ \\
\hline
\end{tabular}
}
\end{center}
\end{table}

\subsection{Results}

Table~\ref{tab:strong} shows that acoustic-only accuracy decreased
monotonically from $0.53$ at $10$~dB to $0.34$ at $-5$~dB, whereas
three-modality accuracy remained between $0.97$ and $0.98$.
In contrast, linguistic-only accuracy remained near the chance level
across all corruption probabilities.
\par
Table~\ref{tab:e2e} shows the same tendency under end-to-end speech
degradation.
CER increased from $0.00$ for clean speech to $0.53$ at $-5$~dB,
while three-modality accuracy remained at $0.98$ throughout.

\subsection{Discussion}

The decrease in acoustic-only accuracy confirms that stronger noise
affects the acoustic representation.
However, three-modality accuracy remains stable because information from
the other modalities compensates for this degradation.
Together with the ablation results in Section~\ref{sec:ablation}, this
shows that the robustness under acoustic variation is not simply a
consequence of weak noise.
\par
The linguistic results differ.
Increasing either synthetic word corruption or end-to-end transcription
errors produces little change because the linguistic modality is already
near the chance level.
Thus, this experiment cannot establish robustness of the linguistic
representation itself.
Rather, it confirms that, with the present linguistic features,
identification using all three modalities is primarily sustained by the
visual and acoustic modalities.

\section{Demonstration on a domestic service robot}
\label{sec:demo}

The project page and demonstration video are
\href{https://okitamisan01.github.io/person_concept/}{available at https://okitamisan01.github.io/person\_concept/}.

\begin{figure*}[t]
\begin{center}
\includegraphics[width=15cm]{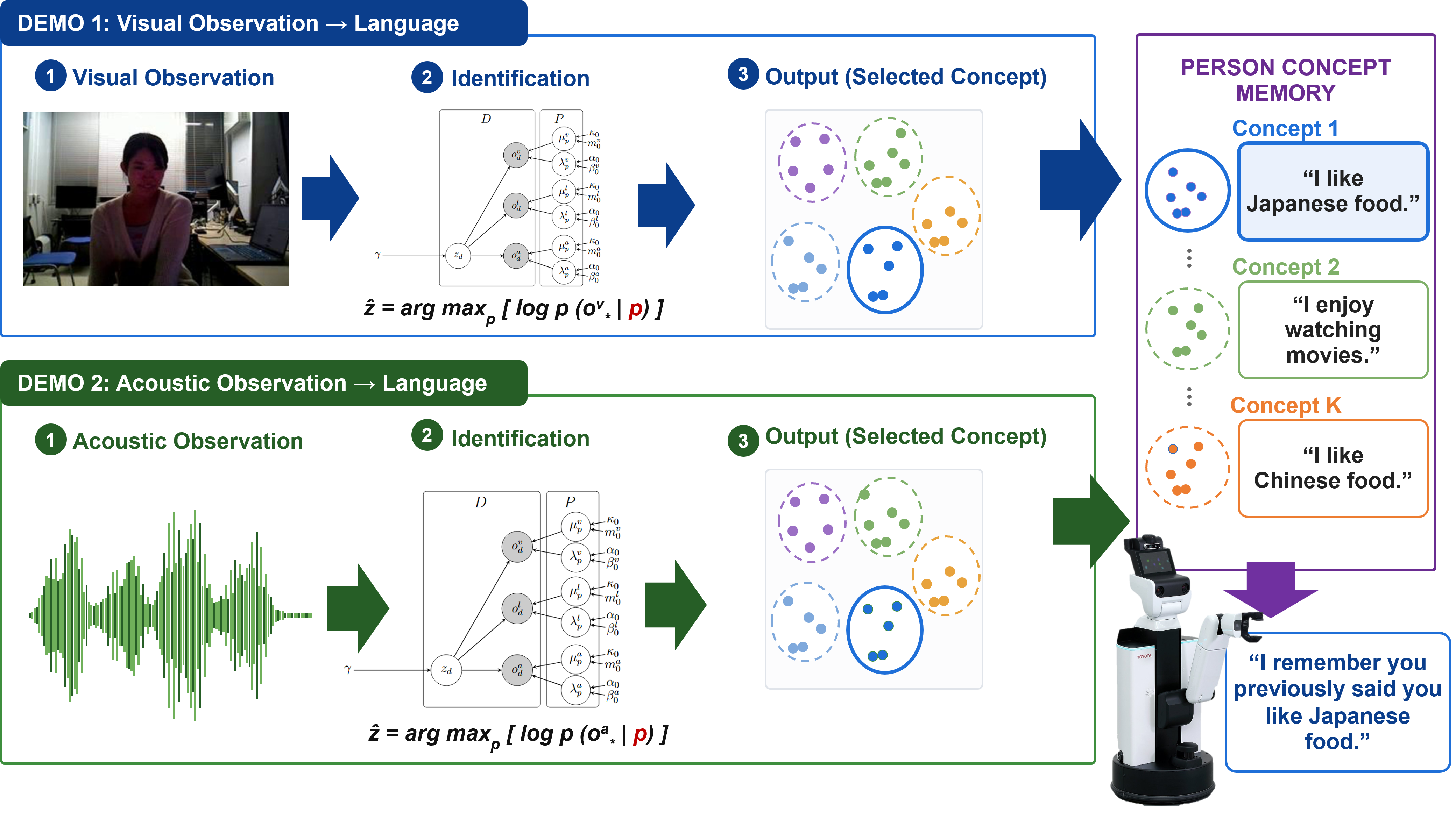}
\caption{\label{fig:demo}
Proof-of-concept demonstration on the HSR.
Demo~1 (top) uses only a visual observation from the HSR camera, and
Demo~2 (bottom) only an acoustic observation.
In each case, the observation is passed to the person concept model,
the concept that maximises the log-likelihood of the available modality
is selected, and the linguistic information stored with the selected
concept is recalled to generate the robot response.
Both modalities accessed the same concept (Concept~1, highlighted),
from which the stored utterance ``I like Japanese food.'' was retrieved.
The person shown in the figure is the first author.}

\end{center}
\end{figure*}

\par
To examine whether person concepts can support person-dependent
interaction on a real robot, we implemented a qualitative demonstration
on a Toyota Human Support Robot (HSR).
As shown in Fig.~\ref{fig:demo}, the HSR accesses a person concept from a single visual or acoustic observation and recalls linguistic information associated with that person.

\par
This demonstration is intended as a qualitative proof of concept rather
than a quantitative evaluation.
In particular, the online configuration differs from that used in the
preceding experiments, as described below.

\subsection{System configuration and concept formation}

The demonstration was implemented with ROS~2.
The HSR acquired visual, acoustic, and linguistic observations, while
feature extraction and concept formation were performed on a PC.
The identification result was displayed on the HSR, and its response was
played through the loudspeaker.
The main configuration is summarized in Table~\ref{tab:demo_setting}.

During observation, the HSR periodically asked questions such as
``What is your favourite food?'' to elicit free speech.
Recording was paused while the HSR spoke, and silence was inserted over
the same interval to preserve the temporal correspondence between the
modalities.
The observations were divided into clips, and CLIP and CLAP features
were extracted.
Person concepts were then formed using the model described in
Section~\ref{sec:model}, and a representative utterance was stored as
linguistic information associated with each concept.
\par
Unlike the quantitative experiments, PCA was not applied because the
observations were acquired incrementally and a fixed projection was not
available in advance.
The original features were therefore used directly.
As shown in Section~\ref{sec:sensitivity}, this high-dimensional setting
can lead to over-segmentation, so the inferred concepts are not expected
to correspond one-to-one with the actual people.
To address this difference, the demonstration focuses on whether a
concept associated with the target person can still be accessed and used
for memory recall, rather than on whether the correct number of concepts
is inferred.

\begin{table}[t]
\caption{\label{tab:demo_setting}
Configuration of the HSR demonstration.}
\begin{center}
{
\renewcommand{\arraystretch}{1.20}
\setlength{\tabcolsep}{5pt}
\begin{tabular}{@{}lc@{}}
\hline
Setting & Value \\
\hline
Participants & $2$ \\
Observation per person & Approx. $3$ min \\
Question interval & Approx. $30$ s \\
Clip length & $10$ s \\
Feature dimension & $512$ \\
Dimensionality reduction & None \\
Identification input & Visual or acoustic \\
Acoustic input for Demo~2 & Approx. $5$ s \\
\hline
\end{tabular}
}
\end{center}
\end{table}

\subsection{Identification and memory recall}

As shown in Fig.~\ref{fig:demo}, two cases were demonstrated.
Demo~1 used only the current visual observation from the HSR camera, whereas Demo~2 used only an acoustic observation.
In each case, the concept was selected from the likelihood of the
available modality; the other modalities were not used for
identification.
The linguistic information stored with the selected concept was then
retrieved and used to generate the HSR response.
In both cases, the concept associated with the target person was
successfully accessed.
For example, after acoustic identification, the HSR recalled a previously
observed food preference and responded,
``I remember you previously said you like Japanese food.''
This was possible even though over-segmentation occurred in the
high-dimensional configuration.
\par
These results qualitatively show that exact estimation of the number of
person concepts is not always required to access person-dependent
information on a robot, and thus that person concepts can support
person-dependent service based on previous interactions.
However, this demonstration does not establish identification or recall
accuracy.
A quantitative evaluation over repeated on-robot trials is left for
future work.

\section{Conclusion}
\label{sec:conclusion}

We proposed a multimodal person concept model that applies
nonparametric multimodal concept formation to person identification.
The model forms person concepts from a small number of visual, acoustic,
and linguistic observations without person labels or prior knowledge of
the number of users.
\par
Through experiments under changing observation conditions, we showed
that person concepts learned from limited observations remain useful for
person identification.
Averaged over the eight conditions, the model was comparable to $k$-means
clustering that was given the true number of participants ($0.89$ versus
$0.88$), although the proposed model inferred the number of person
concepts from the observations alone.
The results also showed that the effect of variation differs across
modalities, particularly for visual observations, and that information
from the remaining modalities can compensate when one modality is
degraded.
\par
The additional experiments further showed that concept formation and
identification should be considered separately.
Feature dimensionality strongly affected the estimation of the number of
concepts, whereas identification remained possible even when the number
of concepts was not correctly estimated.
The modality analysis also showed that, in the present experimental
setting, visual and acoustic observations mainly contributed to concept
formation, while the current English-oriented linguistic representation contributed
little to separating people.
\par
Finally, the HSR demonstration showed that a learned person concept can
be accessed from either a visual or an acoustic observation and used to
recall linguistic information associated with that person.
This supports the feasibility of using learned person concepts for
person-dependent services based on previous interactions.
\par
Future work will evaluate generalisability with more participants,
recording environments, and interaction sessions, using naturally
occurring rather than synthetic variation.
We also plan to investigate sequential person concept formation in real
service environments, such as open-campus events, where a robot
repeatedly encounters different users and accumulates observations over
time.

\textbf{Ethical considerations.}
This study was conducted as an undergraduate research project under the
supervision of the authors' academic advisor. According to guidance
provided by the Human Research Ethics Committee of Soka University,
ethics review was not required for this study under the university's
procedures for undergraduate research.
All participants gave informed consent to the recording and to the use
of the recordings for research.
The only person depicted in this paper is the first author, who appears in
Fig.~\ref{fig:intro}, Table~\ref{tab:dataset}, and Fig.~\ref{fig:demo};
the remaining users in Fig.~\ref{fig:intro} are represented by schematic
silhouettes that do not correspond to any individual.
In the demonstration video, the faces of participants other than the first
author are obscured by mosaic processing.
The example utterances were selected so that they contain no information
identifying a participant.

\end{document}